\documentclass[accepted]{uai2021}
\usepackage[american]{babel}
\usepackage{natbib}
\usepackage{mathtools}
\usepackage{booktabs}
\usepackage{tikz}

\usepackage{multirow}
\usepackage{bm}
\usepackage{amsmath}
\usepackage{hyperref}

\title{TreeBERT: A Tree-Based Pre-Trained Model for Programming Language}

\author[1]{Xue Jiang}
\author[2]{Zhuoran Zheng}
\author[1]{Chen Lyu\textsuperscript{}\thanks{Corresponding author (lvchen@sdnu..edu.cn)}}
\author[1]{Liang Li}
\author[1]{Lei Lyu}

\affil[1]{%
    School of Information Science and Engineering\\
    Shandong Normal University\\
    China
}
\affil[2]{%
     School of Computer Science and Engineering\\
    Nanjing University of Science and Technology\\
    China
}

\begin{document}
\maketitle

\begin{abstract}
Source code can be parsed into the abstract syntax tree (AST) based on defined syntax rules. However, in pre-training, little work has considered the incorporation of tree structure into the learning process. In this paper, we present TreeBERT, a tree-based pre-trained model for improving programming language-oriented generation tasks. To utilize tree structure, TreeBERT represents the AST corresponding to the code as a set of composition paths and introduces node position embedding. The model is trained by tree masked language modeling (TMLM) and node order prediction (NOP) with a hybrid objective. TMLM uses a novel masking strategy designed according to the tree's characteristics to help the model understand the AST and infer the missing semantics of the AST. With NOP, TreeBERT extracts the syntactical structure by learning the order constraints of nodes in AST. We pre-trained TreeBERT on datasets covering multiple programming languages. On code summarization and code documentation tasks, TreeBERT outperforms other pre-trained models and state-of-the-art models designed for these tasks. Furthermore, TreeBERT performs well when transferred to the pre-trained unseen programming language.
\end{abstract}

\section{Introduction}\label{sec:intro}
With the development of pre-trained models such as BERT \citep{Devlin2018}, state-of-the-art performance has been achieved on many natural language processing tasks. The use of pre-trained models reduces the budget of downstream tasks while achieving high accuracy and fast training speed. The success of pre-trained models in natural language processing (NLP) has also driven the emergence of pre-trained models for programming language (PL), such as CuBERT \citep{Kanade2019} and CodeBERT \citep{Feng2020}, which learn generic code representations through pre-training tasks and then use them to enhance PL-oriented downstream tasks, such as code summarization and code classification. 

However, existing PL-oriented pre-trained models face two main challenges.
\textbf{1) Design of appropriate mechanism for learning program syntactical structure.} PL-oriented pre-trained models all adopt a natural language-like treatment, modeling the code as a sequence of words and considering only the linguistic nature of the code. However, the code is also strongly structured, and the semantics of the code relys on the combination of program statements and expressions with different syntactical structures to be represented. Therefore, the structural features of the program cannot be ignored. Typically, a code snippet can be parsed into a unique corresponding abstract syntax tree (AST) based on defined syntax rules. Previous works \citep{DBLP:conf/acl/RabinovichSK17,DBLP:conf/aaai/MouLZWJ16} have noted the key role of AST in program representation. However, the standard Transformer \citep{Vaswani2017} architecture for pre-trained models cannot utilize the tree structure. How to use AST as input to the pre-trained model is a challenging problem.
\textbf{2) Exploration of pre-training tasks for the tree structure.} The existing PL-oriented pre-trained models straightforwardly follow the NLP pre-training tasks. There is some inappropriate application of sequence-oriented tasks directly to non-sequential structured AST. Therefore, new pre-training tasks should be designed for tree so that the pre-trained model can extract both syntactic and semantic information from AST.

In this paper, we propose TreeBERT, \emph{a tree-based pre-trained model for programming language.} TreeBERT follows the Transformer encoder-decoder architecture. To enable the Transformer to utilize the tree structure, we represent the AST corresponding to the code snippet as the set of root node to terminal node paths and then introduce node position embedding to obtain the position of the node in the tree. We propose a hybrid objective applicable to AST to learn syntactic and semantic knowledge, i.e., tree masked language modeling (TMLM) and node order prediction (NOP). In TMLM, we design a novel masking strategy to mask the input of the encoder and decoder, input the set of AST paths with masked nodes at the encoder side, and predict the complete code snippet using the contextual information in AST at the decoder side. Since the order of nodes in the path expresses the program structure information, NOP improves the ability of the model to capture the syntactical structure information by predicting whether the nodes are out of order. As a result, TreeBERT can be applied to a wide range of PL-oriented generation tasks by means of fine-tuning and without extensive modifications to the task-specific architecture.

Our contributions are summarized as follows: 
\begin{itemize}
	\item We propose TreeBERT, a PL-oriented tree-based pre-trained model. TreeBERT is pre-trained on multiple programming language datasets and can be fine-tuned for improving the accuracy of PL-oriented generation tasks. Our source code and model are available at https://github.com/17385/TreeBERT. 
	\item We represent the AST as a set of constituent paths and introduce node position embedding. We propose a hybrid objective for the tree structure, including TMLM and NOP. The former designs a novel masking strategy to help the model understand AST and infer semantics. The latter enables more program syntactical structure features to be extracted.
	\item We conduct extensive experiments on Python, Java, and C\# datasets to verify that TreeBERT achieves state-of-the-art performance in code summarization and code documentation and generalizes better to another programming language that has not been seen in the pre-training phase.
\end{itemize}

\section{Motivation}
In this section, we describe our motivations for proposing TreeBERT, including the program syntactical structure and the pre-training task.

\paragraph{Program syntactical structure}
CuBERT \citep{Kanade2019} represents the first attempt to pre-training a BERT contextual embedding of source code and shows its efficacy on five classification tasks. CodeBERT \citep{Feng2020} is the first bimodal pre-trained model capable of handling programming language (PL) and natural language (NL). It is trained using a hybrid objective function, including standard masked language modeling \citep{Devlin2018} and replacement token detection \citep{Clark2020}, to achieve state-of-the-art performance in NL-PL comprehension tasks (e.g., code retrieval) and generation tasks (e.g., code documentation). Nevertheless, both CuBERT and CodeBERT are based purely on the sequence. The knowledge they learn from the sequence is not sufficient because they do not take into account the learning of syntactical structural information in the program. Additionally, CodeBERT points out that using the code sequence causes a partial loss of accuracy compared to using AST. \citet{Hu2018} use a linear traversal approach to represent AST as the sequence. \citet{Leclair} combine tokens in code with code structure in AST. \citet{Alon2019} represent a given code snippet as the set of paths between k pairs of random terminal nodes in the AST, which makes their model outperform other state-of-the-art techniques in code summarization and code captioning tasks. Each of these task-specific, non-pre-trained models demonstrates that a tree-based representation can effectively extract information from source code than can a sequential representation.

The Transformer \citep{Vaswani2017} has demonstrated strong feature learning capability on NLP tasks, and the strategy of using large amounts of data for pre-training has been successful. Addressing how to enable Transformer to utilize tree structure is the first step in enabling pre-trained models to learn syntactical structure. \citet{DBLP:conf/aaai/SunZXSMZ20} attempt to solve this problem by adding the structural convolution sublayers to the first few Transformer decoder blocks. Instead of changing the Transformer structure, TreeBERT represents the AST using the set of paths from the root to the terminal nodes to serialize the AST. This approach represents a tree uniquely compared to other traversal methods, such as pre-order traversal. We also introduce node position embedding, which obtains hierarchy information and the relative position information of the node's parent and sibling nodes. These operations ensure that the Transformer makes maximum use of all the information in the tree.

\paragraph{Pre-training Task}
Pre-training tasks are crucial for learning generic representations of language. In NLP, masked language modeling \citep{Devlin2018}, next sentence prediction \citep{Devlin2018}, sentence order prediction \citep{Lan2019}, replaced token detection \citep{Clark2020}, permuted language modeling \citep{Yang2019}, and other tasks have been used for pre-training. These specially designed pre-training tasks enable the model to learn contextually relevant representations of each member of the input sentence using a large amount of data, thus the model to implicitly learn generic knowledge in the language.
\begin{figure*}[htb]
	\centering
	\includegraphics[width=1\linewidth]{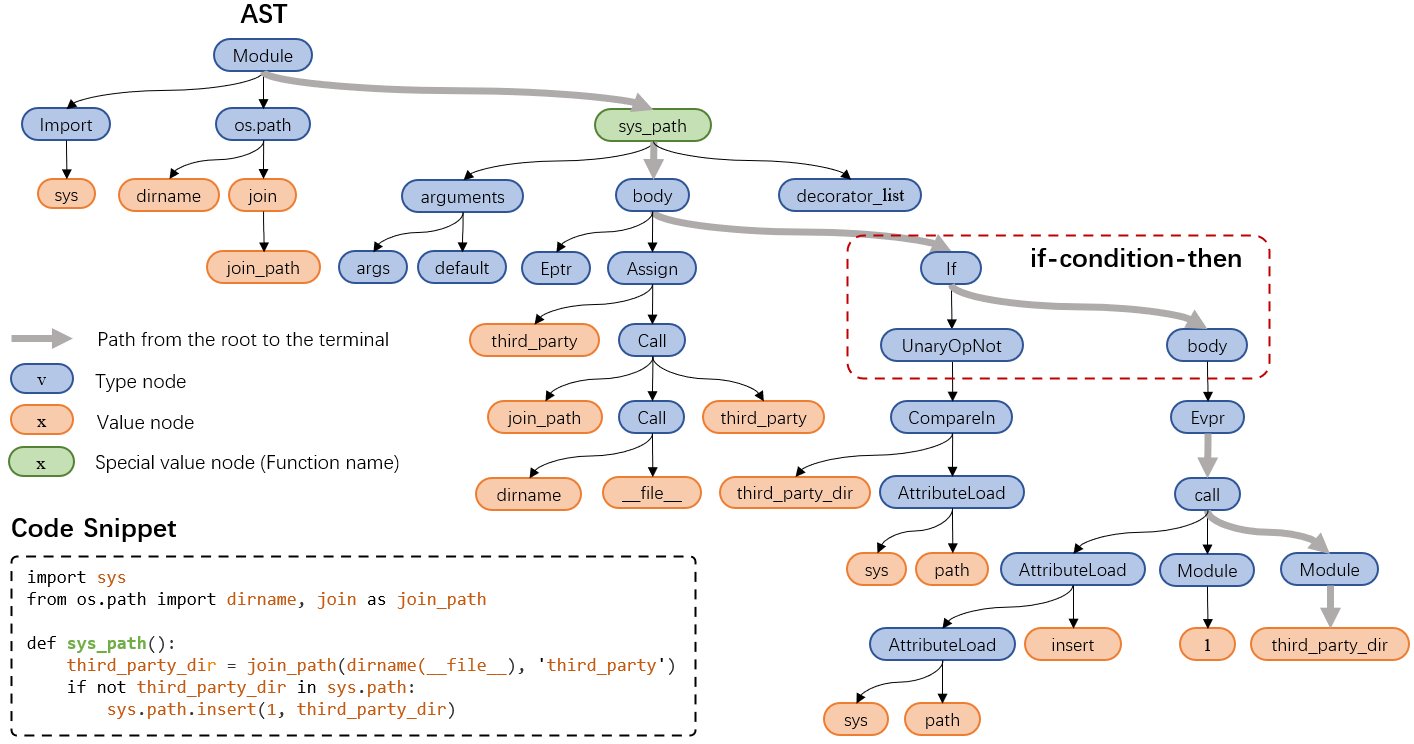}
	\caption{\textbf{AST representation of the code snippet}.  When we represent AST, the terminal node is its corresponding value attribute,  and the non-terminal node is its corresponding type attribute,  except for the function name that acts as a non-terminal node but uses the value attribute.}
	\label{fig:AST}
\end{figure*}

The pre-training tasks of CuBERT and CodeBERT follow the pre-training tasks of NLP. These pre-training tasks are designed for sequence, while non-sequential structured AST has more a complex structure than sequence. It is required to design pre-training tasks for AST to learn the code representation effectively. Our work is based on Seq2Seq MLM used in MASS \citep{Song2019} and T5 \citep{DBLP:journals/jmlr/RaffelSRLNMZLL20} and has been shown to benefit Seq2Seq-style downstream tasks. Specifically, we design TMLM. Considering the repetition of nodes in AST paths and the importance of terminal nodes, we adopt a novel masking strategy to mask the nodes in AST and tokens in code snippet. The input of the encoder is the set of masked AST paths, and the output of the decoder is the predicted complete code snippet, which facilitates TreeBERT to understand AST and infer the missing semantics. In addition, to learn node order constraints in the AST, e.g., the "if" node must be followed by the "body" node, we develop NOP to capture more information about the underlying syntactical structure of the program.

\section{PROPOSED MODEL}
In this section, we describe the details of TreeBERT, including the model architecture, the input representation, and the pre-training tasks employed by TreeBERT.

\subsection{Model Architecture}
The model architecture of TreeBERT is based on the original implementation of the Transformer-based encoder-decoder described by \citet{Vaswani2017}. We omit the detailed description of the Transformer architecture since the use of the Transformer is already widespread \citep{Dong2019,Lan2019,DBLP:conf/acl/LiuHCG19}. Our model modifies the encoder side of the Transformer, adding only a fully connected layer to adjust the dimensionality of the input.

\subsection{Input Representation}
In the pre-training phase, we set the input as a set of constituent paths in the corresponding AST given a code snippet. All node embeddings in the path are concatenated to represent the path after adding the node position embedding. The input at the decoder side is the sequence of tokens obtained from the code snippet after tokenization. In particular, the value nodes in the AST and tokens in the code snippet are represented by summing the subword embeddings.

\paragraph{AST Representation}
Each code snippet is transformed into an AST, as shown in Figure \ref{fig:AST}, which exhibits the syntactical structure of the program in the form of a tree. Each node in the tree represents a structure in the code. For example, a conditional jump statement such as if-condition-then can be represented by a node with two branches. 

AST nodes are divided into two categories: AST type nodes, such as "argument" and "Eptr", denoted as $v$, and AST value nodes (i.e., the tokens in the code), such as "sys\_path" and "join\_path", denoted as $x$. Note that the value nodes are almost always terminal nodes.We use the set of paths from the root node to the terminal nodes to represent the AST, $\mathbf{A}=\{\mathbf{p_1},\mathbf{p_2},\cdots,\mathbf{p_N}\}$, where $N$ is the number of paths contained in the AST.

\paragraph{Code Representation}
The code snippet corresponding to AST is split into a sequence of tokens, and tokens $[LT]$ and $[CLS]$ are added at the beginning and end, respectively, i.e., $\mathbf{C}=\mathbf{[LT]},\mathbf{x_1},\mathbf{x_2},...,\mathbf{x_M },\mathbf{[CLS]}$ where $\mathbf{[LT]}$ is the representation vector of $[LT],$ $\mathbf{[CLS]}$ is the representation vector of $[CLS]$, and $M$ is the length of the code snippet. $\mathbf{C}$ will be used as the input to the decoder, which is shown in Figure \ref{fig:model}. $[EOS]$ is the end-of-sentence identifier at the decoder side. Not only does $[LT]$ serve as a start-of-sentence identifier on the decoder side, but its value represents the type of the target programming language, e.g., $[LT]=[PLT]$ when the language type is Python, $[LT]=[JLT]$ when the language type is Java and $[LT]=[UNK]$ when the language generated by the decoder is a language not seen in the pretraining phase. This is because the implementation details of the different types of languages are hidden when the code snippet is converted to AST. We need to hint at the language type for the model to learn the differences between different programming languages. $[CLS]$ is used as the aggregate representation for NOP.

\paragraph{Path Representation}
Each path is a sequence of nodes, $p_i=v_1^i  v_2^i\cdots v_{L-1}^i  x_t^i$, where the terminal node $x_t^i$ of the path is a token in the corresponding code snippet, and $L$ is the length of the path. We concatenate the vector of nodes in the path to represent the path.
\begin{equation*}\label{eq:example}
	\mathbf{p_i}=Concat[\mathbf{v_1^i};\mathbf{v_2^i};\cdots;\mathbf{v_{L-1}^i};\mathbf{x_t^i}].
\end{equation*}
Note that there is no ordering relationship between path representation vectors in the set of paths. Therefore, unlike the standard Transformer, our model encoder side does not add position encoding to assign position information to the path vectors but adds the position information of the nodes in the tree using node position embedding when forming the node representations.

\paragraph{Token Representation}
A large number of user-defined identifiers in the program often lead to out-of-vocabulary tokens or sparse vector representation. To alleviate this problem, we use bytes-pair-encoding (BPE) \citep{DBLP:conf/acl/SennrichHB16a} to learn the most frequent subtoken from value nodes in the AST and tokens in the code snippet and slice them, e.g., "third\_party" might be sliced into "third", "\_", "party". Following \citet{Alon2019}, we use the vector sum of all subtokens that constitute each token to represent the complete token.
\begin{equation*}
	\mathbf{x_t}=\sum_{s\in BPE\_split(x_t)}E_s^{subtokens},
\end{equation*}
where $E^{subtokens}$ is a learned embedding matrix to represent each subtoken. The number of type nodes in AST is fixed and small, so it is represented as a real-valued vector by embedding.

To train a more general pre-trained model, we share vocabulary and weights for Python and Java. This approach can reduce the overall vocabulary size and maximize the token overlap between languages, thereby improving the cross-language performance of the model \citep{DBLP:conf/nips/ConneauL19,Devlin2018,DBLP:conf/iclr/LampleCDR18}.

\paragraph{Node Position Embedding}
To obtain the relative position information of the nodes in the AST, we add node position embedding to the node embedding on the encoder side.

A node's position embedding is a linear combination of its parent node position embedding and its corresponding level embedding. There are $H+1$ level embeddings as parameters, namely, $W^{level}$, where $H$ is the height of the tree. We use $W_0^{level}$ as the parent position embedding of the root node. If there is a node on level $j$ whose position embedding is $W^{parent}$ and has $c$ child nodes, then the position embedding of its $i$-th child node is
\begin{equation*}
\resizebox{1\hsize}{!}{
	$E_i^{position}\!=\! \frac{c-i+1}{c+1}W^{parent}\!+\!\frac{i}{c+1}W_{j+1}^{level} , c\!\geq\!1 and  1\!\leq i\!\leq\! c,$
	}
\end{equation*}
where $W^{parent}$, $W^{level}$ are learnable weight matrices. Node position embedding can obtain hierarchy information and the relative position information of the node's parent and sibling nodes. 

The pros of node position embedding are fewer parameters (approximately 1/200 of learned position embedding \citep{DBLP:conf/icml/GehringAGYD17} in the experiment) and extrapolability. If n parameters are used, learned position embedding can encode n nodes, while node position embedding is all nodes in a tree of height n-1, the number much larger than n. Learned position embedding, which allocates a different weight matrix as parameters for each position, fails to scale up. Compared to learned position embedding, node position embedding allows the model to extrapolate to a larger number of nodes than the ones encountered during training. 

\subsection{Pre-training Tasks}
\begin{figure}[htb]
	\centering
	\includegraphics[width=1 \linewidth]{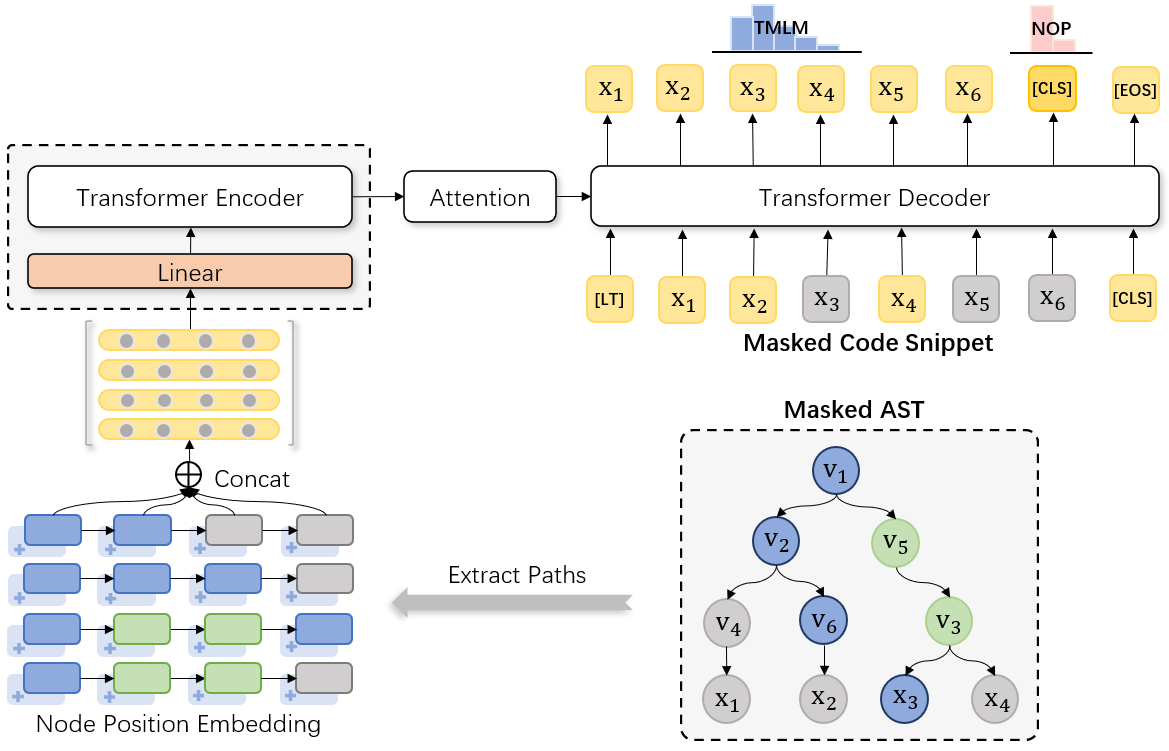}
	\caption{\textbf{Overview of TreeBERT.}\quad The gray nodes indicate that the nodes (or tokens) are masked, and green nodes mean that the nodes (e.g., $v_3$ and $v_5$) exchange their positions.}
	\label{fig:model}
\end{figure}

We design two tasks, TMLM and NOP, for training TreeBERT. Seq2Seq MLM has been shown to be effective in previous studies \citep{Song2019, DBLP:journals/jmlr/RaffelSRLNMZLL20}. We design the TMLM by extending the Seq2Seq MLM to the AST. NOP considers the node order constraints in AST to further extract program structure information. The two pre-training tasks are illustrated in Figure \ref{fig:model}.

\subsubsection{Tree Masked Language Modeling (TMLM)}
Given an AST-code snippet pair ($A$, $C$), we propose a strategy for masking the nodes in the AST and the tokens in the code snippet.

\textbf{\emph{At the encoder side}}, we first sample the nodes in path $p_i$  according to the distribution of probability $\{q_n^i \}_{n=1\cdots L}$ and use the $TOPK()$ operation to select the top $k$ nodes $m_i^A$ with the higher probability and then replace these nodes in path $p_i$ with a special token $[mask]$ to obtain $p_i^{masked}$.
\begin{align*}
	q_n^i & =\frac{e^{\left(l-L\right)}}{\sum_{j=1}^L e^{\left(j-L\right)}},\\
	m_i^A & =TOPK\left(p_i,k,\{q_n^i\}\right),\\
	p_i^{masked} & =REPLACE\left(p_i,m_i^A,[mask]\right),\\
	A^{masked} & =\{p_1^{masked},p_2^{masked},\cdots,p_N^{masked}\},
\end{align*}
where $A^{masked}$ denotes the set of masked paths, $l$ is the current node level, $L$ is the maximum node level in the path, $N$ is the number of paths contained in an AST, and $i=1 \cdots N$. Note that $L$ is subtracted to prevent numerical overflow. This ensures that nodes with larger levels in the path have a higher probability of being masked.

TMLM masks the nodes closer to the terminal in the path with higher probability. The main reasons are 1) since each path is from the root node to the terminal node, the closer the node is to the root node, the more times it is repeated in the set of paths. If we use the standard MLM masking strategy, many same type nodes are masked. Repeatedly learning the representation of such nodes will harm the performance of the model. 2) The terminal nodes of the AST usually refer to user-defined values that represent identifiers and names in the code with rich meanings. Therefore, masking such nodes more frequently can force our model to learn their representations.

\textbf{\emph{At the decoder side}}, the input $C^{masked}$ of the decoder is obtained by masking the tokens in the code snippet according to the following equations:
\begin{align*}
	m^C & =\{x|x\in C \cap x\notin m^A\},\\
	C^{masked} & =REPLACE\left(C,m^C,[mask]\right),
\end{align*}
where $m^A=m_1^A\cup m_2^A \cup \cdots \cup m_N^A$ and $x$ is the element of the set $m^C$ that needs to be masked in the code snippet. We keep the tokens corresponding to the value nodes in $m^A$ and mask the other nodes in the code snippet $C$. In this way, by making the next token prediction, TMLM can force the decoder to rely on the feature representation of the AST instead of the previous tokens in the code snippet.

An example is shown in Figure \ref{fig:model}. In the AST, according to the above strategy, the set of nodes to be masked for the four paths are $m_1^A=\{v_4, x_1\}$, $m_2^A=\{x_2\}$, $m_3^A=\{\}$, and $m_4^A=\{x_4\}$. In the code snippet, the masked value nodes $x_1$, $x_2$, $x_4$ in the path are given, and the other nodes are masked, i.e., $m^C=\{x_3, x_5\}$. The decoder needs to predict the complete code snippet $x_1$, $x_2$, $x_3$, $x_4$, $x_5$, $x_6$.

In TMLM, the encoder reads the set of masked AST paths, and then the decoder infers the code snippet corresponding to the AST. Notably, when the code is converted to AST, some semantic information is hidden, such as "+", ">", "<=" and other binary operators are represented in AST using the "BinOpSub" node. In this case, if the decoder is designed to predict AST, the above semantic information will be ignored. Therefore, we design the decoder to predict the code snippet to encourage the model to infer such semantic information, thus enhancing its generalization ability in downstream tasks. The objective function of TMLM is given as follows:
\begin{equation*}
	\mathcal{L}_{TMLM}\left(\theta\right)=\frac{1}{M}\sum _{x\in C} -log\prod_{t=1}^{M} P\left(x_t | x_{<t} ,A^{masked}\right),
\end{equation*}
where $M$ is the length of the code snippet and $x$ is the token in the code snippet.

In summary, TMLM can force the encoder to understand the AST and infer the semantic information hidden in the AST. It also encourages the decoder to extract information from the encoder to help with code snippet prediction, enabling joint training of the encoder-decoder framework.

\subsubsection{Node Order Prediction (NOP)}
TMLM is capable of learning rich semantic knowledge. To further improve the ability to extract syntactical structure information from program, we designed the binarized pre-training task NOP.

There are some implicit constraints on the order of nodes in the AST. Taking the path "Module ->$\cdots$-> if -> body -> Expr ->$\cdots$-> third\_party\_dir" in Figure \ref{fig:AST} as an example, there must be a "body" node below the "if" node and a "body" node must have an "Expr" node underneath it. To capture this syntactical structure information, we decide with a certain probability whether to randomly exchange the positions of some nodes in the path, and then train the model to distinguish whether the order of nodes in the AST is correct or not. As shown in Figure \ref{fig:model}, we swap the positions of nodes $v_3$ and $v_5$. The hidden vector of $[CLS]$ will be compressed to 1 dimension by a fully connected layer, and then the probability $\bar{y}$ of the existence of nodes out of order in the AST path is obtained via the Sigmoid function. The objective function of NOP is as follows:
\begin{equation*}
	\mathcal{L}_{NOP}\left(\theta\right) = - \left(ylog\bar{y}+\left(1-y\right)log\left(1-\bar{y}\right) \right),
\end{equation*}
where $y$ takes a value of 1 when AST has nodes out of order and is 0 otherwise.

Since TMLM and NOP play different roles, we adopt the hyperparameter $\alpha$ to adjust the loss function.
\begin{equation*}
	\mathcal{L}=\min_\theta \left( \alpha \mathcal{L}_{TMLM}\left(\theta\right) + \left(1-\alpha\right)\mathcal{L}_{NOP}\left(\theta\right)\right).
\end{equation*}

\begin{table*}[htb]
	\centering
	\caption{Experimental results of code summarization on datasets of different sizes in Java and Python.}\label{tab:sum}
	\resizebox{170mm}{25mm}{
		\begin{tabular}{lcccccccccccc}
			\toprule 
			\bfseries  \multirow{2}*{Methods} & \bfseries   & \bfseries ETH Py150& \bfseries & \bfseries   & \bfseries Java-small& \bfseries  & \bfseries   & \bfseries Java-med& \bfseries   & \bfseries   & \bfseries Java-large& \bfseries  \\
			~ & \bfseries Prec & \bfseries Rec& \bfseries F1& \bfseries Prec & \bfseries Rec& \bfseries F1& \bfseries Prec & \bfseries Rec& \bfseries F1& \bfseries Prec & \bfseries Rec& \bfseries F1\\ \midrule 
			\textbf{Methods without pre-training}\\
			Transformer & 22.18 & 12.67& 16.13& 28.13& 26.7& 31.41&50.11& 35.01& 41.22& 59.13& 40.58& 48.13\\
			Graph2Seq & 35.74& 24.77& 29.26& 50.44& 37.15& 42.79& 61.07& 46.89& 53.05& 62.89& 54.07& 58.15\\
			Code2Seq & 36.45& 25.59&30.07&50.64& 37.40& 43.02& 61.24& 47.07&53.23& 64.03& 55.02&59.18\\
			Code+Gnn+GRU & \textbf{40.66}& \textbf{29.89}& \textbf{34.45}& \textbf{55.87}& \textbf{42.32}& \textbf{48.16}& \textbf{67.10}& \textbf{52.35}& \textbf{58.81}& \textbf{71.66}& \textbf{60.56}& \textbf{65.64}\\ \midrule
			\textbf{Methods with pre-training}\\
			MASS(Pre-training with code) & 38.05& 28.14& 32.35& 53.06& 40.15& 45.71& 62.96& 48.56& 54.83& 67.27& 55.45& 60.79\\
			CuBERT & 33.48& 22.61& 26.99& 46.15& 32.62& 38.22& 54.58& 39.73& 45.99& 57.55& 45.07& 50.55\\
			CodeBERT & 35.97& 25.12& 29.58& 49.39& 35.19& 41.10& 58.41& 43.16& 49.64& 61.76& 49.18& 54.76\\\midrule
			TreeBERT & \textbf{45.81}& \textbf{34.01}& \textbf{39.04}& \textbf{60.33}& \textbf{45.68}& \textbf{51.99}& \textbf{69.95}& \textbf{54.42}& \textbf{61.22}& \textbf{73.43}& \textbf{62.03}& \textbf{67.25}\\
			Absolute gain over Code+Gnn+GRU & +5.15& +4.12& +4.58& +4.46& +3.36& +3.83& +2.85& +2.07& +2.40& +1.77& +1.47& +1.61\\
			\bottomrule 
		\end{tabular}
	}
\end{table*}

\section{Experiments}
In this section, we first present the pre-training setup for TreeBERT. Then, we provide the results of fine-tuned TreeBERT on some generation tasks. Finally, an ablation study is performed to evaluate the effectiveness of various components of TreeBERT.

\subsection{Pre-training}
\paragraph{Pre-training Data}
The pre-training datasets we used are the Python and Java corpus published by CuBERT. Duplicate data from the fine-tuned datasets were removed. The Python dataset consists of 7.2 million Python files containing over 2 billion tokens, which are processed into ASTs with approximately 500 million paths and 7 billion nodes. The Java dataset consists of 14.1 million Java files containing approximately 4.5 billion tokens, which are processed into ASTs containing approximately 1.1 billion paths and 16.5 billion nodes.

\paragraph{Training Details}
TreeBERT sets the number of encoder layers (i.e., Transformer blocks) to 6, the hidden size to 1024, and the number of self-attention headers to 8. The decoder uses the same settings as the encoder. For pre-training, the max path number, max node number and max code length are set to 100, 20 and 200, respectively, based on the statistical information about the dataset. We took 32K BPE merge operations. We use Adam \citep{DBLP:journals/corr/KingmaB14} with a learning rate of \{1e-3, 1e-4, 1e-5\}, $\beta1=0.9$, $\beta2=0.999$, $L2$ weight decay of 0.01, learning rate warm up during the first 10\% of steps, and linear decay afterwards. We use a GELU activation, and the batch size is \{2048, 4096, 8192\}. We also applied 0.1 dropout \citep{DBLP:journals/jmlr/SrivastavaHKSS14} on all layers. In TMLM, each path randomly masks 15\% of the nodes. In the NOP, each AST has a 50\% chance of exchanging a pair of nodes.

We train TreeBERT with different values of weight $\alpha$ (0.25, 0.5, 0.75, 1) and test the performance of TreeBERT in two downstream tasks, code summarization and code documentation. The results show that TreeBERT's performance improves as the value of $\alpha$ increases and performs best at 0.75, but the model performance starts to deteriorate when the value of $\alpha$ increases further. Therefore, we set the weight $\alpha$ in the loss function to 0.75.

\subsection{Fine-tuning}
To verify the effectiveness of TreeBERT, TreeBERT was fine-tuned for two generation tasks and was compared with the baselines. The generation tasks are code summarization and code documentation. We also evaluate the performance of TreeBERT on the C\# dataset and experimentally demonstrate that TreeBERT generalizes well to programming language not seen in the pre-training phase.

\subsubsection{Code Summarization}
Code summarization refers to predicting the function name of a given function body, which can roughly describe the function and help programmers name the function.
\paragraph{Datasets and Training Details}
In the code summarization task, we evaluate TreeBERT on Python and Java datasets, where the Python dataset uses ETH Py150 \citep{DBLP:conf/oopsla/RaychevBV16} and the Java datasets \citep{DBLP:conf/icml/AllamanisPS16} use Java-small, Java-med and Java-large. The statistical information of the datasets is provided in the supplementary material.

We set the learning rate and batch size to 1e-5 and 64, respectively. We use the Adam update parameter, and the number of epochs is set to 100. If the model performance does not improve over several epochs, we end the training early and select the model with the lowest loss in the validation set for subsequent evaluation.

\paragraph{Model Comparison}
The selected comparison baselines are the sequence-based Transformer \citep{Vaswani2017}, graph-based Graph2Seq \citep{Xu2018}, AST-based Code2Seq \citep{Alon2019}, and both AST and sequence-based Code+Gnn+GRU \citep{DBLP:conf/iwpc/LeClairHWM20}. Transformer has achieved state-of-the-art performance for translation tasks. Graph2Seq introduces a novel graph-to-sequence neural encoder-decoder model that maps input graphs to vector sequences and decodes target sequences from these vectors using an attention-based LSTM. Code2Seq represents a code snippet as $k$ paths between terminal nodes in its AST and uses attention to select the relevant paths when decoding. Code+Gnn+GRU processes AST using ConvGNN and combines the output of the ConvGNN encoder with the output of the code token encoder. 

We also compared TreeBERT with other pre-trained models, namely, NL-oriented MASS ($L_e=6, L_d=6, H=1024, A=8$) and PL-oriented CuBERT ($L_e=24, H=1024, A=16$) and CodeBERT ($L_e=12, H=768, A=12$), where $L_e$ is the number of encoder layers, $L_d$ is the number of decoder layers, $H$ is the hidden size, and $A$ is the number of self-attention heads. We retrained MASS using the same parameter settings and pre-training data as those used for TreeBERT. In addition, we used the parameters of CuBERT and CodeBERT to initialize Transformer's encoder. In this task, we evaluate the performance of each model using three metrics: precision, recall, and F1 score, which refer to supplementary material for the calculation methods.

Table \ref{tab:sum} lists the results of the code summarization task, and visualizations of the data are shown in the supplementary material. TreeBERT significantly outperformed other models, including both with pre-trained and without pre-trained on all four datasets. Graph2Seq, Code2Seq, and Code+Gnn+GRU outperform Transformer, and these results suggest that tree-based or graph-based representations are more effective than sequence in extracting information from code. MASS, after retraining using the code dataset, outperforms the pre-trained models CuBERT and CodeBERT, suggesting that joint encoder-decoder training improves the generative ability. However, MASS does not perform as well as Code+Gnn+GRU, which was specifically designed for this task. TreeBERT makes more use of the syntactical structure of the code than MASS, and it outperforms Code+Gnn+GRU.

When compared to the best baseline, Code+Gnn+GRU, in terms of the F1 score, TreeBERT performs better on small datasets, where it achieves a relative improvement of 7.95\% on the Java-small dataset. On the smaller ETH Py150 dataset, TreeBERT achieves a relative improvement of 13.29\%. The relative difference becomes smaller as the dataset becomes larger, but it still achieves a relative improvement of 2.45\% on the Java-large dataset. These results indicate that TreeBERT performs better in the absence of annotated data, and it is more robust as the data size increases.

\subsubsection{Code Documentation}
Code documentation refers to generating a comment for the code to understand what the code does by reading the comment rather than the code itself.

\paragraph{Datasets and Training Details}
We use the Java dataset provided by DeepCom \citep{Hu2018} to fine-tune TreeBERT. The size of the original dataset is reduced due to the presence of some code that cannot be converted to AST. The statistics of the processed dataset are provided in the supplementary material.

In this experiment, we set the learning rate and batch size to 3e-5 and 64, respectively, and the Adam optimizer and early stopping are adopted. Finally, we select the best-performing model on the validation set.

\paragraph{Model Comparison}
In the code documentation task, we use the same baseline as for code summarization but add a comparison with DeepCom, which proposes an AST traversal method that converts the AST into the sequence to train a seq2seq model for the code documentation task. As with DeepCom, we adopt BLEU as the metric in this experiment, and its calculation is described in the supplementary material.

\begin{table}[htb]
	\centering
	\caption{Evaluation results of code documentation on Java dataset using BLEU.}\label{tab:doc}
	\setlength{\tabcolsep}{7mm}
	\small
	\begin{tabular}{ll}
		\toprule 
		\bfseries  Methods & \bfseries BLEU\\	\midrule 
		\textbf{Methods without pre-training}\\
		Transformer & 13.32\\
		DeepCom & 14.81\\
		Graph2Seq & 18.39\\
		Code2Seq & 18.48\\
		Code+Gnn+GRU & \textbf{19.73}\\ \midrule
		\textbf{Methods with pre-training}\\
		MASS(Pre-training with code) & 18.92\\
		CuBERT & 17.41\\
		CodeBERT & 17.87\\\midrule
		TreeBERT & \textbf{20.49}\\
		\bottomrule 
	\end{tabular}
\end{table}

Table \ref{tab:doc} shows the results of the code documentation task. TreeBERT achieves a BLEU score of 20.49, which is 7.17 points higher than that of Transformer. In this task, Transformer produces a very poor result because it stays at the shallow textual token-level compared to other methods without pre-training, thereby overlooking information that is important to the program. Transformer is trained from scratch and does not perform as well as methods with pre-training, which proves that pre-trained models learn valuable information. The performance of TreeBERT is significantly improved compared to other baselines because of the joint training of the encoder-decoder using AST.

\subsubsection{Generalization to Programming Language not Seen during Pre-training}
We now evaluate the performance of TreeBERT on programming language not seen in the pre-training phase. We use CodeNN's \citep{DBLP:conf/acl/IyerKCZ16} C\# dataset which contains programming questions and answers about the C\# language obtained from StackOverflow. The dataset statistics are presented in the supplementary material. In this task, we compare TreeBERT with CodeNN, MASS, CuBERT, and CodeBERT. In addition, we evaluate the model using the BLEU metric and the same hyperparameters as those used in code documentation.

\begin{table}[htb]
	\centering
	\caption{Performance of TreeBERT and Baseline models on language not seen in the pre-training phase.}\label{tab:gen}
	\setlength{\tabcolsep}{5mm}
	\small
	\begin{tabular}{ll}
		\toprule 
		\bfseries  Methods & \bfseries BLEU\\ \midrule 
		TreeBERT & \textbf{17.94}\\
		CodeNN & 14.18\\ \midrule
		MASS(Pre-training with code) & \textbf{16.84}\\
		CuBERT & 14.95\\
		CodeBERT & 15.31\\
		\bottomrule 
	\end{tabular}
\end{table}

Table \ref{tab:gen} shows the performance of TreeBERT on C\# dataset. The experimental results indicate that TreeBERT performs generalization surprisingly well on language not seen in the pre-training phase. Specifically, TreeBERT achieves a BLEU of 17.94 by fine-tuning, which is 3.76 points better than that of CodeNN, whose authors introduced the dataset.  We suppose that TreeBERT can be well generalized to the new language because of the high similarity between the AST of the new language and the AST used for pre-training.

\subsection{Ablation Study}
Extensive ablation studies were conducted to better understand the role of the different components and masking strategy in the model. We vary the model in different ways and measure the performance changes for each variant. The experiment is performed on a Java dataset of code documentation task in the following scenarios.

\begin{itemize}
	\item \emph{No TMLM} --- We do not use TMLM.
	\item \emph{No NOP} --- We do not use NOP.
	\item \emph{No Node Position Embedding} --- We use learned position embedding instead of node position embedding.
	\item \emph{Randomly Masking Nodes} --- In TMLM, we do not use our proposed masking strategy and randomly mask nodes in the AST according to the standard MLM.
	\item \emph{Only Masking Value Nodes} --- Since value nodes express rich semantics, we try not to mask type nodes and mask only value nodes.
\end{itemize}
\begin{figure}[htb]
	\centering
	\includegraphics[width=0.98\linewidth]{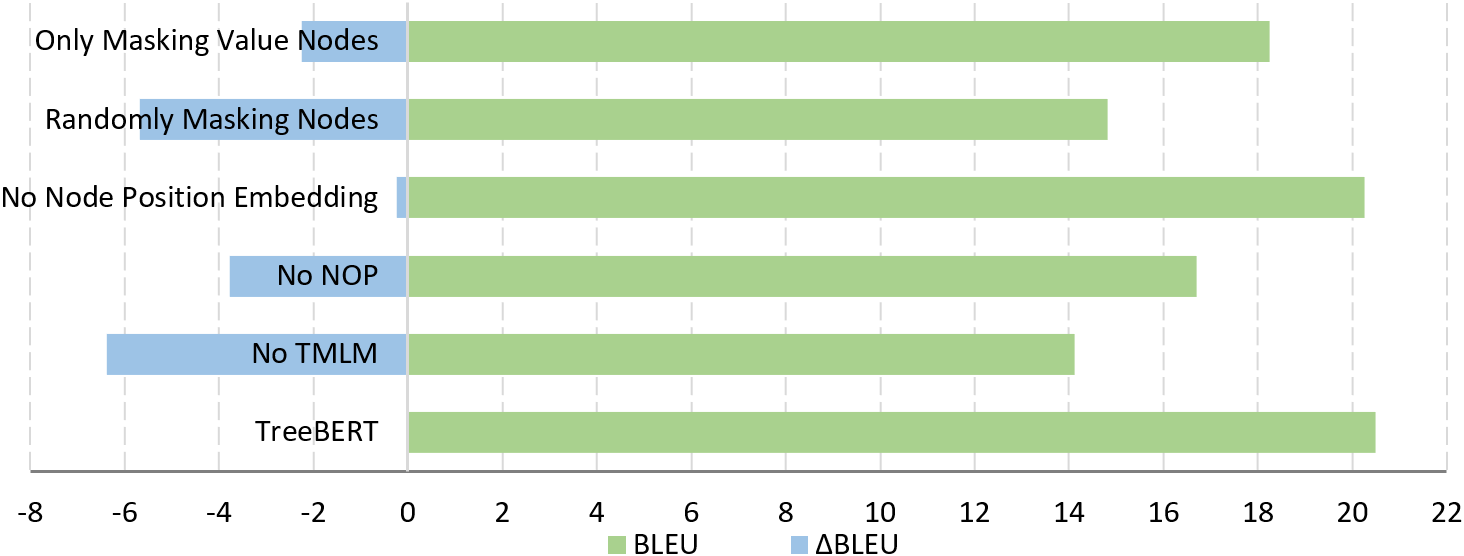}
	\caption{\textbf{Results of ablation study.}\quad The left side shows the loss in BLEU values, and the right side shows the total BLEU values.}
	\label{fig:abla}
\end{figure}
The detailed results of the ablation study are presented in the supplementary material. The experimental results in Figure \ref{fig:abla} show that the model lost the most BLEU when the TMLM is missing, which can prove its important role in TreeBERT. Although the loss is smaller when NOP is missing, the syntactical structure information captured by the NOP is indispensable. When we use node position embedding instead of learning position embedding, the results indicate that we achieve comparable or even higher results than learning position embedding. It is worth mentioning the great role of the masking strategy in TMLM. If the MLM random masking strategy is used, the model will perform poorly, and the result is almost similar to that without TMLM because most of the nodes masked are repeated AST non-terminal type nodes, resulting in a large number of code tokens not being involved in the pre-training, which is detrimental to the extraction of key semantic information in the code. Value nodes represent rich semantics, but some loss can be caused if all masked nodes are value nodes. This suggests that masking a small number of type nodes is useful, and we suppose that this approach can be used to extract structural information and help improve the performance of the model. 

\section{Conclusion}
In this paper, we propose a PL-oriented tree-based pre-trained model called TreeBERT. To learn code representation using AST, TreeBERT uses the set of constituent paths of AST as input, introduces node position embedding and then trains the model using a hybrid objective, including TMLM and NOP. The experimental results indicate that TreeBERT achieves state-of-the-art performance in code summarization and code documentation tasks and performs well when transferred to the pre-training unseen programming language. To further investigate the role of each part of TreeBERT, we conducted an extensive ablation study.

TreeBERT still has considerable room for improvement. First, TreeBERT can be used not only for tasks such as code summarization and code documentation, and it can be used in any task where the source language can be constructed as an AST. We will continue to explore the possibility of applying TreeBERT to more PL tasks. Second, we will further improve TreeBERT, for example, by adding more program information to the AST or by using multimodal forms such as AST, graph and sequence simultaneously, thus extracting information about the program from different perspectives so that TreeBERT can better solve PL downstream tasks.

\begin{acknowledgements}
	This work is financially supported by the National Natural Science Foundation of China (61602286, 61976127) and the Special Project on Innovative Methods (2020IM020100).
\end{acknowledgements}

\bibliography{uai2021-template}

\begin{thebibliography}{29}
\providecommand{\natexlab}[1]{#1}
\providecommand{\url}[1]{\texttt{#1}}
\expandafter\ifx\csname urlstyle\endcsname\relax
  \providecommand{\doi}[1]{doi: #1}\else
  \providecommand{\doi}{doi: \begingroup \urlstyle{rm}\Url}\fi

\bibitem[Allamanis et~al.(2016)Allamanis, Peng, and
  Sutton]{DBLP:conf/icml/AllamanisPS16}
Miltiadis Allamanis, Hao Peng, and Charles Sutton.
\newblock A convolutional attention network for extreme summarization of source
  code.
\newblock In \emph{International Conference on Machine Learning, {ICML}}, 2016.

\bibitem[Alon et~al.(2019)Alon, Brody, Levy, and Yahav]{Alon2019}
Uri Alon, Shaked Brody, Omer Levy, and Eran Yahav.
\newblock code2seq: Generating sequences from structured representations of
  code.
\newblock In \emph{International Conference on Learning Representations,
  {ICLR}}, 2019.

\bibitem[Clark et~al.(2020)Clark, Luong, Le, and Manning]{Clark2020}
Kevin Clark, Minh{-}Thang Luong, Quoc~V. Le, and Christopher~D. Manning.
\newblock {ELECTRA:} pre-training text encoders as discriminators rather than
  generators.
\newblock In \emph{International Conference on Learning Representations,
  {ICLR}}, 2020.

\bibitem[Conneau and Lample(2019)]{DBLP:conf/nips/ConneauL19}
Alexis Conneau and Guillaume Lample.
\newblock Cross-lingual language model pretraining.
\newblock In \emph{Annual Conference on Neural Information Processing Systems,
  NeurIPS}, 2019.

\bibitem[Devlin et~al.(2019)Devlin, Chang, Lee, and Toutanova]{Devlin2018}
Jacob Devlin, Ming{-}Wei Chang, Kenton Lee, and Kristina Toutanova.
\newblock {BERT:} pre-training of deep bidirectional transformers for language
  understanding.
\newblock In \emph{Conference of the North American Chapter of the Association
  for Computational Linguistics: Human Language Technologies, {NAACL-HLT}},
  2019.

\bibitem[Dong et~al.(2019)Dong, Yang, Wang, Wei, Liu, Wang, Gao, Zhou, and
  Hon]{Dong2019}
Li~Dong, Nan Yang, Wenhui Wang, Furu Wei, Xiaodong Liu, Yu~Wang, Jianfeng Gao,
  Ming Zhou, and Hsiao{-}Wuen Hon.
\newblock Unified language model pre-training for natural language
  understanding and generation.
\newblock In \emph{Annual Conference on Neural Information Processing Systems,
  NeurIPS}, 2019.

\bibitem[Feng et~al.(2020)Feng, Guo, Tang, Duan, Feng, Gong, Shou, Qin, Liu,
  Jiang, and Zhou]{Feng2020}
Zhangyin Feng, Daya Guo, Duyu Tang, Nan Duan, Xiaocheng Feng, Ming Gong, Linjun
  Shou, Bing Qin, Ting Liu, Daxin Jiang, and Ming Zhou.
\newblock Codebert: {A} pre-trained model for programming and natural
  languages.
\newblock In \emph{Conference on Empirical Methods in Natural Language
  Processing: Findings, {EMNLP}}, 2020.

\bibitem[Gehring et~al.(2017)Gehring, Auli, Grangier, Yarats, and
  Dauphin]{DBLP:conf/icml/GehringAGYD17}
Jonas Gehring, Michael Auli, David Grangier, Denis Yarats, and Yann~N. Dauphin.
\newblock Convolutional sequence to sequence learning.
\newblock In \emph{International Conference on Machine Learning, {ICML}}, 2017.

\bibitem[Hu et~al.(2018)Hu, Li, Xia, Lo, and Jin]{Hu2018}
Xing Hu, Ge~Li, Xin Xia, David Lo, and Zhi Jin.
\newblock Deep code comment generation.
\newblock In \emph{International Conference on Program Comprehension, {ICPC}},
  2018.

\bibitem[Iyer et~al.(2016)Iyer, Konstas, Cheung, and
  Zettlemoyer]{DBLP:conf/acl/IyerKCZ16}
Srinivasan Iyer, Ioannis Konstas, Alvin Cheung, and Luke Zettlemoyer.
\newblock Summarizing source code using a neural attention model.
\newblock In \emph{Annual Meeting of the Association for Computational
  Linguistics, {ACL}}, 2016.

\bibitem[Kanade et~al.(2020)Kanade, Maniatis, Balakrishnan, and
  Shi]{Kanade2019}
Aditya Kanade, Petros Maniatis, Gogul Balakrishnan, and Kensen Shi.
\newblock Pre-trained contextual embedding of source code.
\newblock 2020.
\newblock URL \url{http://arxiv.org/abs/2001.00059}.

\bibitem[Kingma and Ba(2015)]{DBLP:journals/corr/KingmaB14}
Diederik~P. Kingma and Jimmy Ba.
\newblock Adam: {A} method for stochastic optimization.
\newblock In \emph{International Conference on Learning Representations,
  {ICLR}}, 2015.

\bibitem[Lample et~al.(2018)Lample, Conneau, Denoyer, and
  Ranzato]{DBLP:conf/iclr/LampleCDR18}
Guillaume Lample, Alexis Conneau, Ludovic Denoyer, and Marc'Aurelio Ranzato.
\newblock Unsupervised machine translation using monolingual corpora only.
\newblock In \emph{International Conference on Learning Representations,
  {ICLR}}, 2018.

\bibitem[Lan et~al.(2020)Lan, Chen, Goodman, Gimpel, Sharma, and
  Soricut]{Lan2019}
Zhenzhong Lan, Mingda Chen, Sebastian Goodman, Kevin Gimpel, Piyush Sharma, and
  Radu Soricut.
\newblock {ALBERT:} {A} lite {BERT} for self-supervised learning of language
  representations.
\newblock In \emph{International Conference on Learning Representations,
  {ICLR}}, 2020.

\bibitem[LeClair et~al.(2019)LeClair, Jiang, and McMillan]{Leclair}
Alexander LeClair, Siyuan Jiang, and Collin McMillan.
\newblock A neural model for generating natural language summaries of program
  subroutines.
\newblock In \emph{International Conference on Software Engineering, {ICSE}},
  2019.

\bibitem[LeClair et~al.(2020)LeClair, Haque, Wu, and
  McMillan]{DBLP:conf/iwpc/LeClairHWM20}
Alexander LeClair, Sakib Haque, Lingfei Wu, and Collin McMillan.
\newblock Improved code summarization via a graph neural network.
\newblock In \emph{International Conference on Program Comprehension, ICPC},
  2020.

\bibitem[Liu et~al.(2019)Liu, He, Chen, and Gao]{DBLP:conf/acl/LiuHCG19}
Xiaodong Liu, Pengcheng He, Weizhu Chen, and Jianfeng Gao.
\newblock Annual meeting of the association for computational linguistics,
  {ACL}.
\newblock 2019.

\bibitem[Mou et~al.(2016)Mou, Li, Zhang, Wang, and
  Jin]{DBLP:conf/aaai/MouLZWJ16}
Lili Mou, Ge~Li, Lu~Zhang, Tao Wang, and Zhi Jin.
\newblock Convolutional neural networks over tree structures for programming
  language processing.
\newblock In \emph{{AAAI} Conference on Artificial Intelligence}, 2016.

\bibitem[Rabinovich et~al.(2017)Rabinovich, Stern, and
  Klein]{DBLP:conf/acl/RabinovichSK17}
Maxim Rabinovich, Mitchell Stern, and Dan Klein.
\newblock Abstract syntax networks for code generation and semantic parsing.
\newblock In \emph{Annual Meeting of the Association for Computational
  Linguistics, {ACL}}, 2017.

\bibitem[Raffel et~al.(2020)Raffel, Shazeer, Roberts, Lee, Narang, Matena,
  Zhou, Li, and Liu]{DBLP:journals/jmlr/RaffelSRLNMZLL20}
Colin Raffel, Noam Shazeer, Adam Roberts, Katherine Lee, Sharan Narang, Michael
  Matena, Yanqi Zhou, Wei Li, and Peter~J. Liu.
\newblock Exploring the limits of transfer learning with a unified text-to-text
  transformer.
\newblock \emph{J. Mach. Learn. Res.}, 21:\penalty0 140:1--140:67, 2020.

\bibitem[Raychev et~al.(2016)Raychev, Bielik, and
  Vechev]{DBLP:conf/oopsla/RaychevBV16}
Veselin Raychev, Pavol Bielik, and Martin~T. Vechev.
\newblock Probabilistic model for code with decision trees.
\newblock In \emph{Conference on Object-Oriented Programming Systems,
  Languages, and Applications,{OOPSLA}}, 2016.

\bibitem[Rozi{\`{e}}re et~al.(2020)Rozi{\`{e}}re, Lachaux, Chanussot, and
  Lample]{Lachaux2020}
Baptiste Rozi{\`{e}}re, Marie{-}Anne Lachaux, Lowik Chanussot, and Guillaume
  Lample.
\newblock Unsupervised translation of programming languages.
\newblock In \emph{Annual Conference on Neural Information Processing Systems,
  NeurIPS}, 2020.

\bibitem[Sennrich et~al.(2016)Sennrich, Haddow, and
  Birch]{DBLP:conf/acl/SennrichHB16a}
Rico Sennrich, Barry Haddow, and Alexandra Birch.
\newblock Neural machine translation of rare words with subword units.
\newblock In \emph{Annual Meeting of the Association for Computational
  Linguistics, {ACL}}, 2016.

\bibitem[Song et~al.(2019)Song, Tan, Qin, Lu, and Liu]{Song2019}
Kaitao Song, Xu~Tan, Tao Qin, Jianfeng Lu, and Tie{-}Yan Liu.
\newblock {MASS:} masked sequence to sequence pre-training for language
  generation.
\newblock In \emph{International Conference on Machine Learning, {ICML}}, 2019.

\bibitem[Srivastava et~al.(2014)Srivastava, Hinton, Krizhevsky, Sutskever, and
  Salakhutdinov]{DBLP:journals/jmlr/SrivastavaHKSS14}
Nitish Srivastava, Geoffrey~E. Hinton, Alex Krizhevsky, Ilya Sutskever, and
  Ruslan Salakhutdinov.
\newblock Dropout: a simple way to prevent neural networks from overfitting.
\newblock \emph{J. Mach. Learn. Res.}, 15:\penalty0 1929--1958, 2014.

\bibitem[Sun et~al.(2020)Sun, Zhu, Xiong, Sun, Mou, and
  Zhang]{DBLP:conf/aaai/SunZXSMZ20}
Zeyu Sun, Qihao Zhu, Yingfei Xiong, Yican Sun, Lili Mou, and Lu~Zhang.
\newblock Treegen: {A} tree-based transformer architecture for code generation.
\newblock In \emph{{AAAI} Conference on Artificial Intelligence, {AAAI}}, 2020.

\bibitem[Vaswani et~al.(2017)Vaswani, Shazeer, Parmar, Uszkoreit, Jones, Gomez,
  Kaiser, and Polosukhin]{Vaswani2017}
Ashish Vaswani, Noam Shazeer, Niki Parmar, Jakob Uszkoreit, Llion Jones,
  Aidan~N. Gomez, Lukasz Kaiser, and Illia Polosukhin.
\newblock Attention is all you need.
\newblock In \emph{Annual Conference on Neural Information Processing Systems,
  NeurIPS}, 2017.

\bibitem[Xu et~al.(2018)Xu, Wu, Wang, Feng, and Sheinin]{Xu2018}
Kun Xu, Lingfei Wu, Zhiguo Wang, Yansong Feng, and Vadim Sheinin.
\newblock Graph2seq: Graph to sequence learning with attention-based neural
  networks.
\newblock abs/1804.00823, 2018.
\newblock URL \url{http://arxiv.org/abs/1804.00823}.

\bibitem[Yang et~al.(2019)Yang, Dai, Yang, Carbonell, Salakhutdinov, and
  Le]{Yang2019}
Zhilin Yang, Zihang Dai, Yiming Yang, Jaime~G. Carbonell, Ruslan Salakhutdinov,
  and Quoc~V. Le.
\newblock Xlnet: Generalized autoregressive pretraining for language
  understanding.
\newblock In \emph{Annual Conference on Neural Information Processing Systems,
  NeurIPS}, 2019.

\end{thebibliography}

\appendix

\begin{center}
  \textbf{\Large Supplementary Material}\\[.2cm]
\end{center}

\setcounter{section}{0}
\renewcommand\thesection{\Alph{section}}
\setcounter{figure}{0}
\setcounter{table}{0}
\setcounter{page}{1}
\makeatletter
\renewcommand{\theequation}{S\arabic{equation}}
\renewcommand{\thefigure}{S\arabic{figure}}
\renewcommand{\thetable}{S\arabic{table}}

In this supplemental material, we first introduce the code tokenization in Section \ref{sec:out}. Second, we provide detailed statistical information of datasets used for the experiment in Section \ref{sec:sta}. Then, we describe the metrics used to evaluate TreeBERT in Section \ref{sec:mec}. Finally, we show the detailed results of some experiments in Section \ref{sec:result}.

\section{Code Tokenization}\label{sec:out}
\begin{figure}[htb]
 	\centering
 	\includegraphics[width=0.98\linewidth]{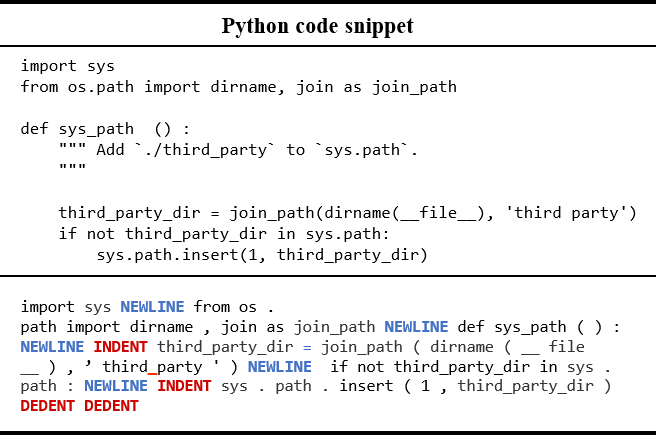}
 	\caption{Example of code tokenization.}
 	\label{fig:out}
 \end{figure}

 Due to the strong structure of code, indentation is meaningful in Python, which cannot be removed simply by splitting code. Follow \citep{Lachaux2020}, we use "INDENT" and "DEDENT" instead of indentation to indicate the beginning and end of a block of code. "NEWLINE" is used to represent line breaks. Spaces are replaced with "\_" in strings, and code comments are removed. An example of a processed Python code snippet is shown in Figure \ref{fig:out}.

\section{Data Statistics}\label{sec:sta}
Table \ref{tab:sumdata} shows detailed statistics of the four datasets used for code summarization, namely, ETH Py150\footnote{\url{https://www.sri.inf.ethz.ch/py150}}, Java-small\footnote{\url{https://s3.amazonaws.com/code2seq/datasets/java-small.tar.gz}}, Java-med\footnote{\url{https://s3.amazonaws.com/code2seq/datasets/java-med.tar.gz}}, and Java-large\footnote{\url{https://s3.amazonaws.com/code2seq/datasets/java-large.tar.gz}}. Table \ref{tab:docdata} shows detailed statistics for two datasets, a Java dataset\footnote{\url{https://github.com/xing-hu/DeepCom/blob/master/data.7z}} from DeepCom \citep{Hu2018} for code documentation and a C\# dataset\footnote{\url{https://github.com/sriniiyer/codenn/tree/master/data/stackoverflow/csharp}} from CodeNN \citep{DBLP:conf/acl/IyerKCZ16} for evaluating the performance of the model on pre-training unseen language.

\begin{table}[h]
	\centering
	\caption{Statistics of datasets used for code summarization.}\label{tab:sumdata}
	\resizebox{84mm}{16mm}{
	\begin{tabular}{lp{1cm}p{1cm}p{1.2cm}p{1.5cm}}
		\toprule 
		\bfseries   & \bfseries  ETH Py150& \bfseries Java-small& \bfseries Java-med & \bfseries Java-large\\
		\midrule 
		Example Number(train) & 143,310 & 665,115& 3,004,536 & 15,344,512\\
		Example Number(valid) & 33,878 & 23,505& 410,699 & 320,866 \\
		Example Number(test) & 35,714 & 56,165& 411,751 & 417,003\\
		Avg.number of Paths(train) & 130 & 171& 187 & 220\\
		Avg.path length(train) & 19 & 21& 23 & 22\\
		Avg.comments length(train) & 3 & 3& 3 & 3\\
		\bottomrule 
	\end{tabular}
}
\end{table}

\begin{table}[h]
	\centering
	\caption{Statistics for DeepCom's Java dataset and CodeNN's C\# dataset.}\label{tab:docdata}
	\begin{tabular}{lll}
		\toprule 
		\bfseries   & \bfseries Java & \bfseries C\#\\
		\midrule 
		Example Number(train) & 450,124 & 52,812\\
		Example Number(valid) & 55,310 & 6,601\\
		Example Number(test) & 54,871 & 6,602\\
		Avg.number of Paths(train) & 212 & 207\\
		Avg.path length(train) & 19 & 16\\
		Avg.comments length(train) & 12 & 10\\
		\bottomrule 
	\end{tabular}
\end{table}

\section{Evaluation Metrics}\label{sec:mec}
In this section, we provide details of the calculation of precision, recall, and F1 score used in the code summarization and BLEU used in code documentation.

\paragraph{Precision, Recall, F1-Score}\label{sec:prf}
In code summarization, we do not use accuracy and BLEU since the generated function names are composed of subtokens and are relatively short (average length of 3 subtokens). Following \citet{Alon2019}., we use precision, recall, and F1 as metrics. The calculation is as follows.
\begin{align*}
	& Precision=\frac{TP}{TP+FP} \\
	& Recall=\frac{TP}{TP+FN} \\
	& F1=\frac{2\cdot Precision \cdot Recall}{Precision + Recall}
\end{align*}
When the predicted subtoken is in the function name, we treat it as a true positive ($TP$). When the predicted subtoken is not in the function name, we treat it as a false positive ($FP$). When the subtoken is in the function name but is not predicted, we treat it as a false negative ($FN$). The label "$UNK$" is counted as $FN$; thus, the prediction of this word will reduce the recall value.

\paragraph{BLEU}\label{sec:bleu}
The BLEU score can be used to measure the similarity between the generated comments and the reference code comments at the sentence level, and it is calculated as follows.
\begin{align*}
	& BLEU=BP \cdot exp \left(\sum_{n=1}^{N} w_n \cdot log p^n\right) \\
	& BP=\left\{
	\begin{aligned}
		1 & , & c \textgreater r, \\
		e^{1-r/c} & , & c \leq r.
	\end{aligned}
	\right.
\end{align*}
where the upper limit of $N$ is taken as 4, i.e., at most 4-grams are computed, $w_n=\frac{1}{N}$, and $p_n$ is ratio of the clauses of length $n$ in the candidate to those also in the reference.

In brevity penalty (BP), $r$ denotes the length of the reference annotation and $c$ denotes the length of the annotation generated by the model.

\section{MORE EXPERIMENTAL RESULTS}\label{sec:result}

\begin{figure}[ht]\label{sec:sunv}
	\centering
	\includegraphics[width=1\linewidth]{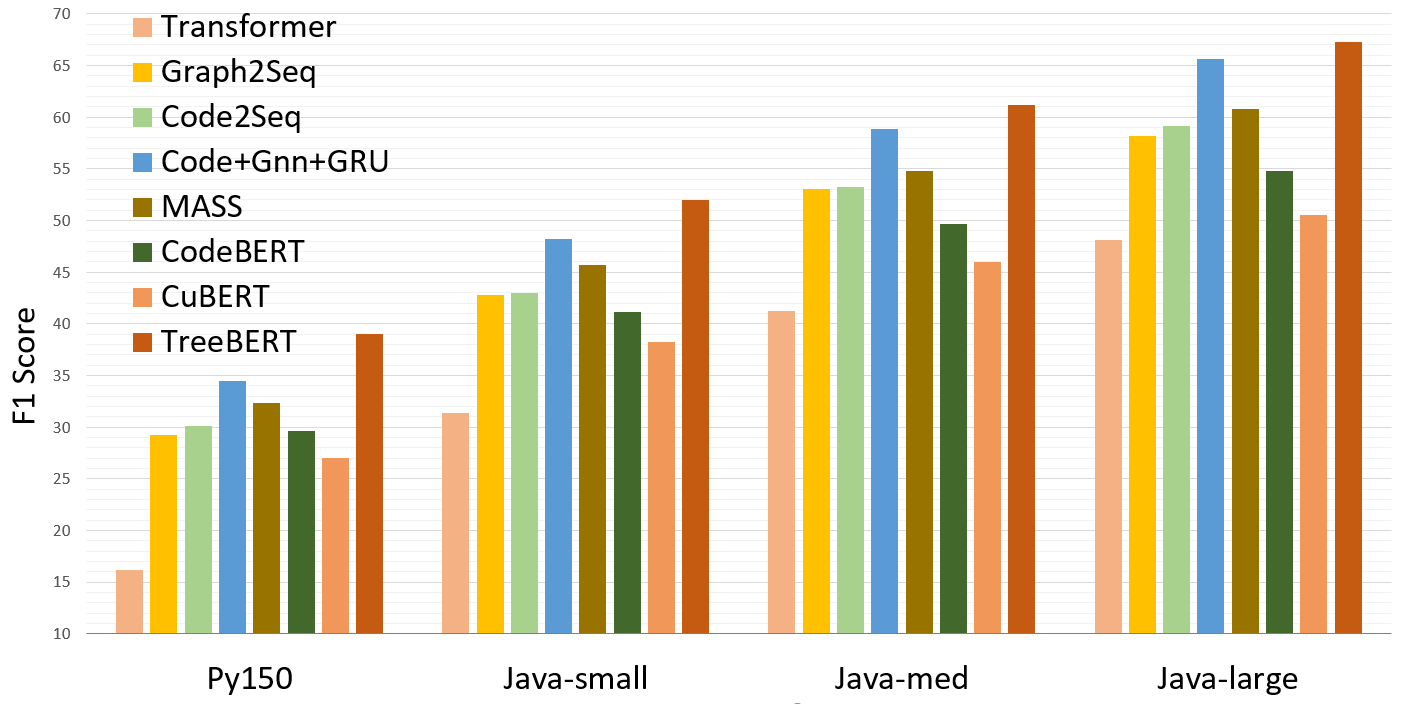}
	\caption{code summarization visualization results for F1 scores on different datasets.}
	\label{fig:sum}
\end{figure}

Figure \ref{fig:sum} shows the visualization results of the F1 score of code summarization. Table \ref{tab:abla} gives the detailed results of the ablation study.

\begin{table}[htb]
	\centering
	\caption{Results of the ablation study.}\label{tab:abla}
	\begin{tabular}{lcc}
		\toprule 
		\bfseries  Model & \bfseries BLEU & \bfseries $\Delta $BLEU\\
		\midrule 
		TreeBERT & 20.49 & -\\
		No PMLM & 14.12 & -6.37\\
		No NOP & 16.71 & -3.78\\
		No Node Position Embedding & 20.25 & -0.24\\
		Randomly Masking Nodes & 14.81 & -5.68\\
		Only Masking Value Nodes & 18.25 & -2.24\\
		\bottomrule 
	\end{tabular}
\end{table}

\end{document}